\documentclass[runningheads]{llncs}

\usepackage{graphicx}
\usepackage{xcolor}
\usepackage{amssymb}
\usepackage{tabularx}
\usepackage{hyperref}

\usepackage{listings}
\usepackage{color, colortbl}
\definecolor{gray_1}{RGB}{142,152,178}
\definecolor{mygray}{gray}{0.95}
\usepackage[normalem]{ulem}
\newcommand{\old}[1]{\textcolor{red}{\ifmmode\text{\sout{\ensuremath{#1}}}\else\sout{#1}\fi}}

\newcommand{\ltilde}{\raisebox{0.7ex}{\texttildelow}}

\begin{document}
%--------------------------------------------------------------------------------------------------
\title{Automated Reasoning in Non-classical Logics \\
in the TPTP World\thanks{%
The first and second authors acknowledge financial support from the Luxembourg National Research 
Fund (FNR), under grant CORE C20/IS/14616644.
The third author acknowledges financial support from the German Federal Ministry for Economic 
Affairs and Energy within the project “KI Wissen – Entwicklung von Methoden für die Einbindung 
von Wissen in maschinelles Lernen", project number 19A20020J.}}

\titlerunning{Non-classical Logics in the TPTP World}

\author{
Alexander~Steen\inst{1,2}\orcidID{0000-0001-8781-9462} \and \\
David~Fuenmayor\inst{2,6}\orcidID{0000-0002-0042-4538} \and \\
Tobias~Glei{\ss}ner\inst{3}\orcidID{0000-0002-7730-5852} \and \\
Geoff~Sutcliffe\inst{4}\orcidID{0000-0001-9120-3927} \and \\
Christoph~Benzm{\"u}ller\inst{5,6}\orcidID{0000-0002-3392-30935}}

\authorrunning{A. Steen et al.}

\institute{
University of Greifswald, Germany \and
Universit{\'e} du Luxembourg, Luxembourg \and
Fraunhofer FOKUS, Germany \and
University of Miami, USA \and
University of Bamberg, Germany \and
Freie Universit{\"a}t Berlin, Germany}

\maketitle              % typeset the header of the contribution
\begin{abstract}
Non-classical logics are used in a wide spectrum of disciplines, including artificial intelligence,
computer science, mathematics, and philosophy.
The de-facto standard infrastructure for automated theorem proving, the TPTP World, currently 
supports only classical logics. 
Similar standards for non-classical logic reasoning do not exist (yet).
This hampers practical development of reasoning systems, and limits their interoperability and 
application.
This paper describes the latest extension of the TPTP World, which provides languages and
infrastructure for reasoning in non-classical logics.
The extensions integrate seamlessly with the existing TPTP World.

\end{abstract}
%--------------------------------------------------------------------------------------------------

\section{Introduction}
\label{Introduction}

The TPTP World \cite{Sut17} is a well established infrastructure that supports research, 
development, and deployment of Automated Theorem Proving (ATP) systems.
The TPTP World includes the TPTP problem library,
% \cite{Sut09}, 
the TSTP solution library,
% \cite{Sut10}, 
standards for writing ATP problems and reporting ATP solutions,
% \cite{SS+06,Sut08-KEAPPA}, 
tools and services for processing ATP problems and solutions,
% \cite{Sut10}, 
and it supports the CADE ATP System Competition (CASC).
% \cite{Sut16}.
Various parts of the TPTP World have been deployed in a range of applications,
in both academia and industry.
The web page \href{http://www.tptp.org}{\texttt{http://www.tptp.org}} provides access to all 
components.

The TPTP languages are one of the keys to the TPTP World's success.
The languages are used for writing both TPTP problems and TSTP solutions, which enables convenient 
communication between different systems and researchers. It also enables tool exchange, tool 
integration, and comparable experimental results.
Originally the TPTP World supported only first-order clause normal form (CNF) \cite{SS98-JAR}.
Over the years full first-order form (FOF) \cite{Sut09}, typed-first order form (TFF) 
\cite{SS+12,BP13-TFF1}, and typed higher-order form (THF) \cite{SB10,KSR16} have been added.
The TFF and THF languages include constructs for arithmetic.

This paper describes the latest extension of the TPTP World, which provides languages and
infrastructure for reasoning in non-classical logics \cite{Pri08,Gob01}, via the (new)
typed extended first-order non-classical (TXN) and typed higher-order non-classical (THN) 
languages. 
TXN and THN support a broad range of non-classical logics.
The formulae of the problem/solutions, and also the specific logic to be used for reasoning, are 
expressed in the same language framework.
In this paper we exemplify the languages using modal logics \cite{BBW06}.
%, e.g., alethic modal logic \cite{Men15}.
However, at all times the reader should keep in mind that the languages do have much broader 
capability.
For example, syntactically, the new languages allow multiple non-classical logics to be used 
together (while, of course, the semantic implications of using such combinations need to be 
carefully considered).
In the medium to long term it is hoped that experts in various non-classical logics will
use the TPTP framework to develop specifications that can be assimilated into the TPTP World.
Stakeholders are invited to contribute!\footnote{%
Send email to the fourth author, {\tt geoff@tptp.org}.}

\paragraph{Motivation.}
The development of standards for ATP systems for first- and higher-order logic has traditionally 
focused mostly on classical logic, while many real-world applications often also require 
non-classical reasoning. 
These include, for example, topical applications in artificial intelligence (e.g., knowledge 
representation, planning, and multi-agents systems), philosophy (e.g., formal ethics and 
metaphysics), natural language semantics (e.g., generalized quantifiers and modalities), and 
computer science (e.g., software and hardware verification).
There are also recent developments in natural and life sciences that employ logical reasoning
(e.g., modelling of biochemical processes).

There has been a gradual disconnect between classical and non-classical logics in the practical 
development and handling of automated reasoning technology, with classical logics receiving
greater attention.
This is unfortunate because there exist quite effective ATP systems for different non-classical 
logics, but their usage, interoperability, and incorporation within larger contexts is hampered 
by their heterogeneous input formats and non-uniform modes of result reporting.
Furthermore, various non-classical logics can be reduced to classical logics, e.g., the
well-known standard translation of modal logics to first-order logic \cite{Ohl93}.
As a consequence many classical ATP systems have not yet been fully evaluated for their potential 
for non-classical reasoning modulo such translation, as their experimental results
cannot be (easily) compared to results from specialized non-classical ATP systems.

This work provides a fruitful bridge between the different communities, and, in particular, fosters
the interoperability and comparability of classical and non-classical reasoning systems.
A preliminary format proposal was discussed in earlier work~\cite{WSB16}.

\paragraph{Related work.}

The ``DFG syntax''~\cite{HKW96}, a format for problem and proof interchange developed in the 
DFG Schwerpunktprogramm Deduktion, contains a meta-information tag called {\em logic} 
%in the description part of the file, 
that can be used to specify ``non-standard quantifiers or operators'' in informal natural language.
This has, however, up to the authors' knowledge, not been used actively.
% because the logic description 
%itself is given in informal natural language.

The Knowledge Interchange Format (KIF)~\cite{GF92} is a comprehensive format for knowledge 
representation, including numbers, lists, sets, and non-monotonic rules. 
KIF could be considered a language for non-classical logic. 
However, KIF is based on a first-order language and comes with a fixed semantics. 
It is not flexible enough to capture different logics.

Common Logic (CL) is an ISO standard~\cite{II18} for the representation of logical information,
with several dialects and a common general XML-based syntax. 
While allowing expressing both first-order and higher-order concepts, it also comes with a fixed 
semantics.% and does not offer flexible representation of domain-specific non-classical logics.

The OMDoc format~\cite{Koh06} is also XML-based, and is geared primarily towards uniform 
representation of mathematical knowledge.
Related to the latter is MMT~\cite{KR16}, which covers and heavily redesigns the formal subset 
OMDoc. MMT aims at providing foundation independent means of specifying formal systems.

%--------------------------------------------------------------------------------------------------
\section{The TPTP Languages}
\label{TPTPLanguages}

The TPTP languages are human-readable, machine-parsable, flexible and extensible languages,
suitable for writing both ATP problems and ATP solutions.
The new TPTP languages described in this paper support the representation of ATP problems and 
solutions in non-classical logics.
In this section the general structure of the TPTP languages is reviewed, and key features of
the TXF and THF languages that underlie the new non-classical languages are presented.
The syntax of the TPTP languages is available in an extended BNF \cite{VS06} at
\href{http://www.tptp.org/TPTP/SyntaxBNF.html}{\texttt{http://www.tptp.org/TPTP/SyntaxBNF.html}}.

%--------------------------------------------------------------------------------------------------
\subsection{The Structure of the TPTP Languages}
\label{Structure}

The top-level building blocks of the TPTP languages are {\em annotated formulae}.
An annotated formula has the form:\\
\hspace*{1cm}{\em language}{\tt (}{\em name}{\tt ,} {\em role}{\tt ,} {\em formula}{\tt ,} 
{\em source}{\tt ,} {\em useful\_info}{\tt ).} \\ 
where the {\em source} and {\em useful\_info} are optional.
The {\em language}s supported are clause normal form ({\tt cnf}), first-order form ({\tt fof}), 
typed first-order form ({\tt tff}), and typed higher-order form ({\tt thf}).
The {\em role}, e.g., {\tt axiom}, {\tt lemma}, {\tt conjecture}, defines the use of the formula 
in an ATP system.
In the {\em formula}, terms and atoms follow Prolog conventions.
The TPTP language also supports interpreted symbols, which either start with a {\tt \$}, 
e.g., the truth constants {\tt \$true} and {\tt \$false}, or are composed of non-alphanumeric 
characters, e.g., numbers (see Section~\ref{TFFTHF}).
The basic logical connectives are {\tt !}, {\tt ?}, {\tt \ltilde}, 
{\tt |}, {\tt \&}, {\tt =>}, {\tt <=}, {\tt <=>}, and {\tt <\ltilde>}, 
for $\forall$, $\exists$, $\neg$, $\vee$, $\wedge$, $\Rightarrow$, $\Leftarrow$, $\Leftrightarrow$, 
and $\oplus$ (xor) respectively. 
Equality and inequality are expressed as the infix operators {\tt =} and {\tt !=}.
The {\em source} and {\em useful\_info} are optional information about the origin and useful
details of the formula.
An example annotated first-order form (FOF) formula defining the set-theoretic union operation, 
supplied from a file named {\tt SET006+1.ax}, is~\ldots
\[
\begin{minipage}{\textwidth}
\begin{verbatim}
    fof(union,axiom,
        ( ! [X,A,B] :
            ( member(X,union(A,B))
          <=> ( member(X,A) | member(X,B) ) ),
        file('SET006+0.ax',union),
        [description('Definition of union'), relevance(0.9)]).
\end{verbatim}
\end{minipage}
\]

%--------------------------------------------------------------------------------------------------
\subsection{The Existing TFF and THF Languages}
\label{TFFTHF}

The typed first-order form (TFF) language extends the first-order form (FOF) language with types 
and type declarations.
Predicate and function symbols can be declared before their use, with type signatures that 
specify the types of the symbols' arguments and results.
The defined types are {\tt \$tType} for the ``type of types'' (used for declaring new base types),
{\tt \$o} for the Boolean truth-values, {\tt \$i} for individuals, {\tt \$int} for integers, 
{\tt \$rat} for rationals, and {\tt \$real} for reals.
In TFF, the expression {\tt ($t_1\,*\ldots*\,t_n$)$\,$>$\,$\$o} declares the type of an $n$-ary 
predicate, where the $i$-th parameter is of type $t_i$ and it returns a Boolean.
Types of functions are analogously declared as {\tt ($t_1\,*\ldots*\,t_n)\,$>$\,t$},
for an $n$-ary function that returns a term of type $t$.
TFF supports arithmetic with numeric constants such as 27, 43/92, -99.66, and 
arithmetic predicates and functions such as {\tt \$greater} and {\tt \$sum}.
A useful feature of TFF is default typing for symbols that are not declared:
predicates default to {\tt (\$i\,*\ldots*\,\$i)\,>\,\$o}, and
functions default to {\tt (\$i\,*\ldots*\,\$i)\,>\,\$i}.
This allows TFF to effectively degenerate to untyped FOF.

The monomorphic variant of TFF is called TF0.
For example~\ldots 

\[
\begin{minipage}{\textwidth}
\begin{verbatim}
    tff(dog_type,type,      dog: $tType ).
    tff(human_type,type,    human: $tType ).
    tff(owner_of_decl,type, owner_of: dog > human ).
    tff(bites_decl,type,    bites: (dog * human * $int) > $o ).
    tff(hates_decl,type,    hates: (human * human) > $o ).
    
    tff(dog_bites_human_more_than_once,axiom,
        ! [D: dog,H: human,N: $int] :
          ( ( bites(D,H,N)
            & $greater(N,1) )
         => hates(H,owner_of(D)) ) ).
\end{verbatim}
\end{minipage}
\]

The typed higher-order form (THF) extends the typed first-order form with higher-order 
notions, including adoption of curried form for type declarations, lambda terms with a lambda 
binder {\tt \verb|^|} for $\lambda$, application with {\tt @}, a choice binder {\tt @+} for 
$\epsilon$, and a description binder {\tt @-} for $\iota$.
In THF all symbols must be declared before their use (default typing is not possible).

The monomorphic variant of THF is called TH0.
For example~\ldots

\[
\begin{minipage}{\textwidth}
\begin{verbatim}
    thf(fix_decl,type,fix: ($o > $o) > $o > $o ).
    thf(fix_defn,definition, 
        fix = (^ [F: $o > $o, X: $o] : ( (F @ X) = X )) ).
    
    thf(id,conjecture,
        ! [F: $o > $o] : 
          ( (! [X: $o] : (fix @ F @ X) ) 
         => F = (^ [X: $o] : X) ) ).
\end{verbatim}
\end{minipage}
\]

The polymorphic extensions of TFF and THF, called TF1 \cite{BP13-TFF1} and TH1 \cite{KSR16},
add type constructors, type variables, and polymorphic symbols.
TH1 also adds five polymorphic constants:
{\tt !!} for $\Pi$, {\tt ??} for $\Sigma$, {\tt @@+} for $\epsilon$, {\tt @@-} for $\iota$,
and  {\tt @=}  for typed equality.
The monomorphic subsets TF0 and TH0 are currently more widespread than the polymorphic extensions, 
and they are the basis for the non-classical languages introduced in this paper.

%--------------------------------------------------------------------------------------------------
\subsection{The Extended TXF and THF Languages}
\label{TXFTHF}

Since the introduction of TFF and THF there have been some features that have 
received little attention: tuples, conditional expressions (if-then-else), and let 
expressions (let-defn-in).
Recently conditional expressions and let expressions have become more important because of their 
use in software verification \cite{KKV18}.
In a separate development, Evgeny Kotelnikov et al. \cite{KKV15} introduced the FOOL logic that 
extends TFF so that (i)~formulae of type \texttt{\$o} can be used as terms, (ii)~variables of type 
\texttt{\$o} can be used as formulae, (iii)~tuple terms and tuple types
are available as first-class citizens, and (iv)~conditional and let expressions are supported.
This logic can be automatically translated to first-order logic \cite{KKV15}.

%--------------------------------------------------------------------------------------------------
\paragraph{The Extended TXF Language.}
%\label{TXF}

The typed extended first-order form (TXF) \cite{SK18}\footnote{%
The language was called TFX in \cite{SK18}, and has now been renamed to TXF.} 
is a superset of the TFF language, including all the features of FOOL logic. 
TXF provides the basis for the typed extended first-order non-classical (TXN) language described 
in Section~\ref{TXNTHN}.
The features of TXF that are most relevant to TXN are Boolean terms and variables: 

\begin{itemize}
  \item Formulae of type \texttt{\$o} can be used as terms, e.g.,~\ldots
\begin{verbatim}
    tff(p_decl,type,  p: ( $i * $o * $int ) > $o ).
    tff(q_decl,type,  q: ( $int * $i ) > $o ).
    tff(me_decl,type, me: $i ).
    tff(fool_1,axiom,
        ! [X: $int] : p(me, ! [Y: $i] : q(X,Y), 27) ).
\end{verbatim}
  \item Variables of type \texttt{\$o} can be used as formulae, e.g.,~\ldots
\begin{verbatim}
    tff(implies_decl,type, implies: ( $o * $o ) > $o ).
    tff(implies_defn,definition,
        ! [X: $o,Y: $o] : ( implies(X,Y) <=> (~(X) | (Y)) ) ).
\end{verbatim}
\end{itemize}

%--------------------------------------------------------------------------------------------------
\paragraph{The (not really extended) THF Language.}
%\label{THF}

In parallel to the development of TXF, THF has been revised to have the same structures as TXF 
for tuples, conditional expressions, and let expressions. 
The revised THF provides the basis for the typed higher-order non-classical (THN) language 
described in Section~\ref{TXNTHN}.
In THF the features of FOOL are naturally available, and thus their presentation in the TXF
context is immediately adopted in THF.
Examples corresponding to those above for TXF are~\ldots

\[
\begin{minipage}{\textwidth}
\begin{verbatim}
    thf(p_decl,type, p: $i > $o > $int > $o ).
    thf(q_decl,type, q: $int > $i > $o ).
    thf(me_decl,type, me: $i ).
    thf(fool_1,axiom,
        ! [X: $int] : ( p @ me @ (! [Y: $i] : (q @ X @ Y)) @ 27 ) ).
\end{verbatim}
\end{minipage}
\]

\ldots\ and \ldots

\[
\begin{minipage}{\textwidth}
\begin{verbatim}
    thf(implies_decl,type, implies: $o > $o > $o ).
    thf(implies_defn,definition,
        ! [X: $o,Y: $o] : ( (implies @ X @ Y) <=> (~ X | Y) ) ).
\end{verbatim}
\end{minipage}
\]

%--------------------------------------------------------------------------------------------------
\section{The TXN and THN Languages}
\label{TXNTHN}

The typed extended first-order non-classical (TXN) and typed higher-order non-classical (THN) 
languages are the new TPTP languages for non-classical logics, extending TXF and THF respectively
(note the mnemonic `{\em N}' in the names TXN and THN, indicating {\em N}on-classical).
The design of TXN and THN adopted the following principles:
(i)~syntactic consistency with the underlying classical languages,
(ii)~a uniform syntax for a wide range of non-classical logics, and
(iii)~requiring minimal changes to existing parsing and reasoning software.
The new languages add new non-classical connectives, in a long form (Section~\ref{ConnectivesLong})
and in some cases a short form (Section~\ref{ConnectivesShort}), and a syntax for specifying
the logic that should be used for reasoning (Section~\ref{Specifications}).
% to allow formulae as arguments of the non-classical connectives.
The salient extract from the full TPTP language grammar is provided in Appendix~\ref{Grammar}.

%--------------------------------------------------------------------------------------------------
\subsection{The Non-Classical Connectives (Long Form)}
\label{ConnectivesLong}

TXN and THN add a new interpreted functor-like connective form~\ldots\\
\hspace*{1cm}{\tt \verb|{|}{\em connective\_name}{\tt \verb|}|} \\
The {\em connective\_name} is a TPTP defined symbol or system symbol, i.e., starting with {\tt \$} 
or {\tt \$\$}, naming a non-classical connective.
If the {\em connective\_name} is a TPTP defined symbol then its meaning is documented in the 
TPTP.
If the {\em connective\_name} is a system symbol then its meaning is defined by the 
user/ATP system being used, thus allowing the TPTP syntax to be used when experimenting with 
logics that have not been formalized in the TPTP.
A {\em connective\_name} may optionally be parameterized, as explained below.
In TXN the non-classical connectives are applied in first-order functional style, while in THN 
their application is explicit~\ldots
\begin{itemize}
\item In TXN {\tt \verb|{|}{\em connective\_name}{\tt \verb|}|}{\tt (}{\em arg$_1$}{\tt ,}{\em ...}{\tt ,}{\em arg$_n$}{\tt )}
      is a formula, where each {\em arg$_i$} is a TXN term.
      TXN terms are defined as for TXF, which permits formulae as arguments, and nesting of
      formulae.
\item In THN {\tt \verb|{|}{\em connective\_name}{\tt \verb|}|} {\tt @} {\em arg$_1$} {\tt @} {\em ...} {\tt @} {\em arg$_n$}
      is a formula, where each {\em arg$_i$} is a THN term.
      THN terms are defined as for THF, which permits nesting of formulae.
\end{itemize}

\noindent
Some examples using connectives from modal logic are~\ldots

\[
\begin{minipage}{\textwidth}
\begin{verbatim}
    tff(pigs_fly_decl,type,pigs_fly: $o ).
    tff(flying_pigs_impossible,axiom,
        ~ {$possible}(pigs_fly) ).
    tff(something_is_necessary,axiom,
        ? [P: $o] : {$necessary}(P) ).

    thf(positive_decl,type,positive: ($i > $o) > $o ).
    thf(self_identity_is_positive,axiom,
        {$necessary} @ (positive @ ^[X:$i] : (X = X)) ).
    thf(everything_is_possibly_positive,axiom,
        ! [P: $i > $o] : ({$possible} @ (positive @ P)) ).
\end{verbatim}
\end{minipage}
\]
Note that the language identifiers {\tt tff} and {\tt thf} are retained from the underlying 
classical languages, which minimizes the amount of adaptation necessary in existing parsers, etc.

A {\em connective\_name} may optionally be parameterized to reflect more complex 
non-classical connectives, e.g., in multi-modals logics where the modal operators are indexed,
in epistemic logics~\cite{vDH15} where the common knowledge operator can specify the agents 
under consideration, and in dynamic logics \cite{HKT00} where the connectives are parameterized 
with (complex) programs.
The form is~\ldots \\
\hspace*{1cm}{\tt \verb|{|}{\em connective\_name}{\tt (}{\em param$_1$}{\tt ,}{\em \ldots}{\tt ,}{\em param$_n$}{\tt )}{\tt \verb|}|} \\
If the connective is indexed, i.e., representing a family of connectives parameterized over some 
index set of constants, the index is given as the first argument as a constant (uninterpreted 
constant, number, or TPTP defined constant) prefixed with a {\tt \#}.
All other parameters are key-value pairs of the form~\ldots\\
\hspace*{1cm}{\em parameter\_name} {\tt :=} {\em parameter\_value} \\
where the {\em parameter\_name} is a TPTP defined symbol or a system symbol, i.e., starting 
with {\tt \$} or {\tt \$\$}, and the {\em parameter\_value} is any term.
In many logics, including the examples from modal logics below, the parameter values 
(including index values) are on the meta level.
They are thus distinct from symbols (even of the same name) occurring at the object level, and 
are not declared with types. 
In the future more complex logics such as term-modal logics~\cite{FTV01} or term-sequence 
modal logics~\cite{SSY19} might merge these levels; the syntax does not prohibit this.

Some parameterized examples using connectives from epistemic logics are given below, where 
{\tt \$knows(\#agent)} is the knowledge operator for {\tt agent}, and {\tt \$common} is the 
common knowledge operator for a set of agents encoded as a key-value parameter 
{\tt \$agents:=\verb|[|\ldots\verb|]|}~\ldots
\[
\begin{minipage}{\textwidth}
\begin{verbatim}
    tff(pigs_fly_decl,type,pigs_fly: $o ).
    tff(alice_knows_pigs_dont_fly,axiom,
        {$knows(#alice)}(~ pigs_fly) ).
    tff(abc_know_pigs_dont_fly,axiom,
        {$common($agents:=[alice,bob,claire])}(~ pigs_fly) ).

    thf(positive_decl,type,positive: ($i > $o) > $o ).
    thf(alice_and_bob_know_self_identity_is_positive,axiom,
        {$common($agents:=[alice,bob])} @ 
          (positive @ ^ [X:$i] : (X = X)) ).
    thf(everything_is_known_to_alice,axiom,
        ! [P: $o] : ( {$knows(#alice)} @ P ) ).
\end{verbatim}
\end{minipage}
\]

As was noted in Section~\ref{TFFTHF}, the default typing rules of TFF and TXF, and hence 
also of TXN, allows them to degenerate to untyped languages.
In the following example \texttt{bird} and \texttt{fly} default to predicates of type
\texttt{\$i > \$o}, \texttt{tweety} defaults to a constant of type \texttt{\$i}, and \texttt{X}
defaults to a variable of type \texttt{\$i}. 
It uses an exemplary system-defined non-classical binary connective {\tt \$\$usually}, denoting 
some kind of (not further specified) non-monotonic conditional.
\[
\begin{minipage}{\textwidth}
\begin{verbatim}
    tff(birds_fly,axiom,
        ! [X] : {$$usually}(bird(X),fly(X)) ).
    tff(tweety_is_bird,axiom, bird(tweety) ).
    tff(tweety,conjecture, fly(tweety) ).
\end{verbatim}
\end{minipage}
\]

%--------------------------------------------------------------------------------------------------
\subsection{The Short Form Connectives}
\label{ConnectivesShort}

For any given logic, some of the long form connectives can be associated with short forms
that provide a more lightweight representation. 
There are three short forms: {\tt [\ldots]}, {\tt <\ldots>}, and {\tt /\ldots\verb|\|},
of which only some might be used in any particular logic.
The \ldots\ can be a {\tt \#}-prefixed index, or a single dot to indicate no parameters.
Key-value pairs cannot be used in the short forms, but long and short forms can be used together.
Short forms for non-classical connectives are documented in the TPTP, or defined by the 
user in case of system defined connectives.
For example, in alethic modal logic 
{\tt [.]} is short for (unindexed) {\tt \verb|{|\$necessary\verb|}|},
{\tt <.>} is short for (unindexed) {\tt \verb|{|\$possible\verb|}|},
{\tt [\#index]} is short for {\tt \verb|{|\$necessary(\#index)\verb|}|} and 
{\tt <\#index>} is short for {\tt \verb|{|\$possible(\#index)\verb|}|}.
Examples corresponding to the long form examples above are~\ldots
\[
\begin{minipage}{\textwidth}
\begin{verbatim}
    tff(pigs_fly_decl,type,pigs_fly: $o ).
    tff(flying_pigs_impossible,axiom,
        ~ <.>(pigs_fly) ).
    tff(something_is_necessary,axiom,
        ? [P: $o] : [.](P) ).

    thf(positive_decl,type,positive: ($i > $o) > $o ).
    thf(self_identity_is_positive,axiom,
        [.] @ (positive @ ^[X:$i] : (X = X) ) ).
    thf(everything_is_possibly_positive,axiom,
        ! [P: $i > $o] : (<.> @ (positive @ P)) ).
\end{verbatim}
\end{minipage}
\]

In epistemic logic {\tt /\ldots\verb|\|} is the short form for {\tt \$believes}.
Examples aligned with the long form examples above (but not those involving common knowledge that
require key-value parameters, which are not allowed in short forms) are~\ldots
\[
\begin{minipage}{\textwidth}
\begin{verbatim}
    tff(pigs_fly_decl,type,pigs_fly: $o ).
    tff(alice_believes_pigs_dont_fly,axiom,
        /#alice\(~pigs_fly) ).

    thf(everything_is_believed_by_alice,axiom,
        ! [P: $o] : ( /#alice\ @ P ) ).
\end{verbatim}
\end{minipage}
\]

%--------------------------------------------------------------------------------------------------
\section{Logic Specifications}
\label{Specifications}

In the world of classical logics, the intended (classical) logic can often be inferred from the 
language used for the formulae. 
This is not so in the world of non-classical logics where
the same language can used for formulae while different logics are used for reasoning.
A well known example is provided by the modal logic cube \cite{Gar18}, which shows the hierarchical 
relationships between different modal logics that use the same language.
When reasoning, e.g., about metaphysical necessity, modal logic \textbf{S5} is usually used, but 
when reasoning about deontic necessities a more suitable choice might be modal logic \textbf{D}. 
Thus when a formula uses a modal connective, e.g., $\Box$ ({\tt \{\$necessary\}} or 
{\tt \verb|[.]|} in TXN/THN syntax),
it is unknown what notion of necessity is intended, and in quantified logics it is unknown how 
necessity interacts with quantification, e.g., whether $\forall x.\,\Box P(x)$ entails 
$\Box \forall x.\,P(x)$ (in TXN syntax, whether {\tt !\,[X:\,\$i]\,:\,\verb|[.]|(p(X))} entails
{\tt \verb|[.]|(!\,[X:\,\$i]\,:\,p(X))}.
It is therefore necessary to provide a new kind of \mbox{(meta-)}information that specifies the 
logic to be used.
A new kind of TPTP annotated formula has been introduced for this, with the role \texttt{logic},
and a ``logic specification'' as the formula.

A logic specification consists of a defined logic name identified with a list of properties, e.g., 
in TXF~\ldots\\
\hspace*{1cm}{\tt tff(}{\em name}{\tt,logic,}{\em logic\_name} {\tt ==} {\em properties}{\tt ).} \\
where {\em properties} is a {\tt \verb|[]|}ed list of key-value identities~\ldots\\
\hspace*{1cm}{\em property\_name} {\tt ==} {\em property\_value} \\
where each {\em property\_name} is a TPTP defined symbol or a system symbol,
and each {\em property\_value} is either a term of the language (often a defined constant) or a 
{\tt \verb|[]|}ed list that might start with a term (often a defined constant), and otherwise 
contains key-value identities.
If the first element of a {\em property\_value} is a term then that is the default value for all 
cases that are not specified by the following key-value identities.
The parameter names and values for a growing number of logics will documented in the TPTP.
A simple example from modal logic is~\ldots
\[
\begin{minipage}{\textwidth}
\begin{verbatim}
   tff(simple_spec,logic, 
       $modal == [
         $constants == $rigid,
         $quantification == [ $constant, human_type == $varying ],
         $modalities == $modal_system_S5 ] ).
\end{verbatim}
\end{minipage}
\]
See Section~\ref{Examples} for more sophisticated examples, and 
Section~\ref{SyntaxOfLogicSpecifications} in particular for the explanation of the
{\em property\_name}s and {\em property\_value}s used here.
The salient extract from the TPTP language BNF is provided in Appendix~\ref{Grammar}.

The grammar is quite unrestrictive, which allows it to be used for many different logics. 
The grammar allows quite complicated specifications, e.g., arbitrary formulae can be used 
as {\em property\_value}s.
It is also possible to specify the same logic in different ways, 
for users to create specifications
for logics that are not defined in the TPTP, and 
(users beware) to write meaningless specifications in a syntactically well-formed way! 
A tool to check the sanity of a specification is available (see Section~\ref{Tools}).

The logic specification typically comes first in a TPTP problem file, and it is an error to use 
a non-classical connective without a logic specification.
The logic specification binds meta-logical information to the object-level information in the 
problem formulae.
Note that the logic specification can change the meaning of language features such as 
truth-values, universal quantification, etc. -- existing meanings from classical logic should 
not be confused with the meanings in the declared logic. 

%--------------------------------------------------------------------------------------------------
\section{Case Study: Multi-Modal Logics}
\label{Examples}

Quantified normal multi-modal logics \cite{BBW06} is the first family of non-classical logics 
defined in the TPTP. 
The standardization originates from preliminary work \cite{GSB17,GS18} that has subsequently been
translated to the format described in this paper.
In this section the TPTP representation of (quantified) multi-modal logics is introduced, and 
the logic specification parameters are discussed. 
A logic puzzle is presented to exemplify its usage.

%--------------------------------------------------------------------------------------------------
\subsection{Syntax and Logic Specification}
\label{SyntaxOfLogicSpecifications}

The formula language of quantified normal multi-modal logics is that of classical logics without 
equality (i.e., all classical connectives are available as usual), augmented with a unary 
connective $\Box$, with an indexed form $\Box_i$.
The reading of $\Box \varphi$ depends on the application context, such as 
``$\varphi$ {\em is necessary}'', ``$\varphi$ {\em is obligatory}'', and 
``$\varphi$ {\em is known}''.
From here forward these connectives are used without any assumption about the intended reading 
unless stated.
The dual $\Diamond$ (and similarly $\Diamond_i$) is defined by 
$\Diamond \varphi := \neg \Box \neg \varphi$.
Note that any multi-modal language can also be regarded a mono-modal language if there is only 
one index value.

Quantified normal multi-modal logic is named {\tt \$modal} in the TPTP.
The connectives are named {\tt \verb|{|\$box\verb|}|} and {\tt \verb|{|\$dia\verb|}|}, with the 
indexed forms {\tt \verb|{|\$box(i)\verb|}|} and {\tt \verb|{|\$dia(i)\verb|}|}.
The corresponding unindexed short forms are {\tt \verb|[.]|} and {\tt <.>}, and the indexed
short forms are {\tt \verb|[|\#i\verb|]|} and {\tt <\#i>}.
The index identifiers are uninterpreted constants on the meta-level, as described in 
Section~\ref{ConnectivesLong}.
For increased readability, the TPTP also defines specialized modal logics with more specific 
names for the connectives.
The logics are \verb|$alethic_modal|, \verb|$deontic_modal|, and \verb|$epistemic_modal|. 
Each of the these is identical to \verb|$modal| in terms of syntax and parameterization except
that {\tt \verb|{|\$box\verb|}|} and {\tt \verb|{|\$dia\verb|}|} are renamed to
\verb|{$necessary}| and \verb|{$possible}|, \verb|{$obligatory}| and \verb|{$permissible}|, and 
\verb|{$knows}|, respectively.

Logics specifications for {\tt \$modal} use three semantically oriented parameters that
characterize the logic to be used.
The parameter names and their possible values are shown in Table~\ref{modal_parameters}.\footnote{%
It is important to note that the parameters for \texttt{\$modal} could have also been chosen 
to characterize the logic via proof-theoretic properties. 
As an example, the parameter \texttt{\$quantification} can equivalently replaced by 
parameters that indicate whether the (converse) Barcan formula is a tautology or not.}

\begin{table}[hbt]
\caption{Logic specification parameters of \texttt{\$modal}.}
\label{modal_parameters}
\centering
\begin{tabular}{p{.22\textwidth}|p{.75\textwidth}}
{\bf Parameter} & {\bf Valid values} \\[.2em]
\hline
{\tt \$constants}      & {\tt \$rigid}, {\tt \$flexible} \\[0.2em]
\rowcolor{mygray}
{\tt \$quantification} & {\tt \$constant}, {\tt \$varying}, {\tt \$cumulative}, {\tt \$decreasing} \\[0.2em]
{\tt \$modalities}     & {\tt \$modal\_system\_X} \\
                       & {\tt X} $\in$ \{{\tt K}, {\tt KB}, {\tt K4}, {\tt K5}, {\tt K45}, 
                         {\tt KB5}, {\tt D}, {\tt DB}, {\tt D4}, {\tt D5}, {\tt D45}, {\tt T},
                         {\tt B}, {\tt S4}, {\tt S5}, {\tt S5U}\} \\
                       & {\em or a list of axioms} \\
                       & {\tt [\$modal\_axiom\_X$_1$, \$modal\_axiom\_X$_2$, ...]} \\
                       & {\tt X$_i$} $\in$ \{{\tt K}, {\tt T}, {\tt B}, {\tt D}, {\tt 4}, {\tt 5}, 
                         {\tt CD}, {\tt BoxM}, {\tt C4}, {\tt C}\} \\
\hline
\end{tabular}
\end{table}

\begin{itemize}
\item The \verb|$constants| parameter specifies whether (function) symbols are interpreted 
      as {\tt \$rigid}, i.e. interpreted as the same domain element in every world, or as
      {\tt \$flexible}, i.e., possibly interpreted as different domain elements in different 
      worlds.
      The parameter can provide a value for all symbols, or a default value and individual 
      values for some symbols.
\item The \verb|$quantification| parameter specifies restrictions on the quantification domain
      across the accessibility relation \cite{FM98}, with the possible values {\tt \$constant}, 
      {\tt \$varying}, {\tt \$cumulative}, and {\tt \$decreasing}.
      The parameter can provide a value for all types, or a default value and individual 
      values for some types.
\item The \verb|$modalities| parameter specifies properties of the connectives.
      Possible values are defined for well-known modal logic systems, 
      e.g.,~\texttt{\$modal\_system\_K}, and individual modal axiom schemes, e.g., 
      \texttt{\$modal\_axiom\_5}.
      They refer to the corresponding systems and axiom schemes from the modal logic cube
      \cite{Gar18}.
      The parameter can provide a value for all indices, or a default value and individual 
      values for some indices.
\end{itemize}

An example (a more sophisticated version of the example in Section~\ref{Specifications}) is ...
\begin{verbatim}
    tff(complex_spec,logic, 
        $modal == [
          $constants ==      [ $flexible, 
                               sun == $rigid ],
          $quantification == [ $constant,
                               planet_type == $varying,
                               human_type == $varying ],
          $modalities ==     [ $modal_system_K, 
                               [#1] == $modal_system_KB,
                               [#2] == [ $modal_axiom_K, 
                                         $modal_axiom_4 ] ] ).
\end{verbatim}
In this example: $\bullet$~all symbols are flexible except for the symbol {\tt sun} that is rigid,
$\bullet$~quantification over variables of any type is over a constant domain, except for 
variables of type {\tt planet\_type} and {\tt human\_type} that are over varying domains, and
$\bullet$~the default modality is system {\tt K}, but index {\tt \#1} uses system {\tt KB} and
index {\tt \#2} uses the axiom schemes {\tt K} and {\tt 4}.

The TPTP provides multiple roles to distinguish between various types of formulae that are assumed
to be true at the start of reasoning, e.g., {\tt axiom}, {\tt hypothesis}, {\tt lemma}, etc.;
these are collectively referred to as {\em axiom-like} roles.
Following the generalized notion of consequence by Fitting and Mendelsohn~\cite{FM98}, in 
{\tt \$modal} the role \texttt{hypothesis} is used to indicate that the formula is assumed to be 
true {\em locally}, i.e., in the current world, and all other axiom-like roles, e.g., {\tt axiom}, 
{\tt lemma}, etc., are used to indicate that the formula is assumed to be true {\em globally}, 
i.e., in all worlds.
Conjectures are, by default, to be proven locally, i.e., in the current world.
In the distant past the TPTP supported {\em subroles} that were to be used to further specify 
the intended meaning of roles.
This feature is being revived to allow the local/global defaults to be overridden, e.g., 
a formula with the role {\tt axiom-local} will be local, and a conjecture with the role 
{\tt conjecture-global} will be global.

%--------------------------------------------------------------------------------------------------
\subsection{Application Example}
\label{ApplicationExample}

Four non-classical logicians, Tim, Fred, Betty and Nancy, walked into a bar.\footnote{%
They were probably proponents of the FDE logic \cite{Bel92-4V}: 
Tim ordered a whisky ({\em true}), Fred ordered a glass of water ({\em false}), 
Betty ordered {\em both}, and Nancy ordered {\em neither}. 
But that's just a humorous coincidence that does not impact this example.} 
They form the steering committee (SC) of a non-classical logic conference. 
As the night goes on, and the empty glasses pile up, they start discussing the conference bylaws.
Since one of the agreed rules is that all SC decisions are made by majority vote, they start
arguing about the following (quite reasonable) rule:

 ``The number of SC members is necessarily an odd number.''

\noindent 
The situation is formalized in TXN as follows (\texttt{eq} represents an adequately axiomatized 
equality predicate)~\ldots
\[
\begin{minipage}{\textwidth}
\begin{verbatim}
    tff(four_members,hypothesis, eq(scMemberCount,4) ).
    tff(four_not_odd,hypothesis, ~ odd(4) ).
    tff(agreed_rule, hypothesis, {$necessary}(odd(scMemberCount)) ).
\end{verbatim}
\end{minipage}
\]
\noindent The discussion goes on as follows:

Tim: \textit{This rule is hopelessly inconsistent: 4 is not an odd number. 
It cannot possibly be! Let's better forget about it.}

Fred: \textit{I disagree, the rule {\em per se} is not inconsistent. 
The reason is that you take the term ``the number of SC members'' to {\em rigidly} denote the 
number 4.}

Tim's assumption that constants denote rigidly is represented in a TXN logic specification 
by~\ldots
\[
\begin{minipage}{\textwidth}
\begin{verbatim}
    tff(tim,logic, 
        $modal ==
          [ $constants == $rigid, 
            $quantification == $constant,
            $modalities == $modal_system_S5 ] ).
\end{verbatim}
\end{minipage}
\]
In this setting the state of affairs is indeed inconsistent as, e.g., confirmed by Leo-III
(see Appendix~\ref{DetailedProblem} for a complete presentation of the problem):
\begin{verbatim}
    % No. of inferences in proof: 22
    % SZS status Unsatisfiable for puzzle.tim.p
\end{verbatim}

Fred continues: \textit{A better alternative is to take 4 as {\em flexibly} denoting 
whatever number of SC members there happen to be. 
So if we were, say, 3 SC members, the rule would be perfectly fine. 
But I agree with you that, right now, the rule is of no use for us, since we can derive a 
contradiction from it, namely, that 4 is an odd number, so anything would follow \ldots}

Unlike Tim, Fred reasons assumes that \texttt{scMemberCount} denotes flexibly. 
However, he also employs an (alethic) modal logic that assumes necessity implies truth, i.e.,
adopting the modal axiom $T$ ($\Box A \rightarrow A$)~\ldots
\[
\begin{minipage}{\textwidth}
\begin{verbatim}
    tff(fred,logic, 
        $modal ==
          [ $constants == [ $rigid, scMemberCount == $flexible ],
            $quantification == $constant,
            $modalities == [ $modal_axiom_K, $modal_axiom_T ] ] ).
\end{verbatim}
\end{minipage}
\]
Betty: \textit{I agree with interpreting the term ``the number of SC members'' flexibly as you 
suggest. 
However, I don't see the rule deriving a contradiction. 
That something is necessarily the case does not imply that something is actually the case.
So the number of SC members is necessarily odd, yet it is four in the actual world. 
I don't see any trouble with this!}

Betty assumes \texttt{scMemberCount} denotes flexibly, while using a modal logic that does not 
assume the modal axiom $T$.
For instance, this can be the modal logic system \textbf{D} (aka.~\textit{standard deontic 
logic} -- SDL, where $\Box$ is read as ``obligatory'')~\ldots
\[
\begin{minipage}{\textwidth}
\begin{verbatim}
    tff(betty,logic, 
      $modal ==
        [ $constants == [ $rigid, scMemberCount == $flexible ],
          $quantification == $constant,
          $modalities == $modal_system_D ] ).
\end{verbatim}
\end{minipage}
\]

Nancy: \textit{Yes, I agree. 
The rule is perfectly consistent and, moreover, we should adopt it now! 
However, this means that we are actually violating the rule, so either someone else must come 
or one of us must go!} she says, looking at Tim.

Nancy also assumes that constants denote flexibly, while employing a more sophisticated logic, 
e.g., the deontic system \textbf{E} \cite{Aqv84,BFP19}.
In contrast to SDL this logic is suitably extended to deal with norm violations (e.g., 
\textit{contrary-to-duty} reasoning) so that they do not result in inconsistencies. 
Alas, such a logic is not easily captured in {\tt \$modal}, and it might be necessarily
to a more expressive logic employing a different specification.

Tim: \textit{But we have to decide this by majority vote!}

%--------------------------------------------------------------------------------------------------
\section{Tools for the TPTP}
\label{Tools}

The TPTP problem library v9.0.0 will include modal logic problems in the TXN and THN languages.
It is expected to be released in the second half of 2022 or the first half of 2023.
The TPTP4X utility \cite{Sut07-CSR} will include output formats for existing non-classical ATP
systems, to provide a bridge to the TPTP problems until they adopt a TPTP language natively. 
Contemporary ATP systems to bridge to include, e.g., K\raisebox{-3pt}{S}P~\cite{NHD20,PN+21}, 
nanoCoP 2.0 \cite{Ott21}, MleanCoP \cite{Ott14}, MetTeL2 \cite{TSK12}, LoTREC \cite{FF+01}
and MSPASS \cite{HS00-TABLEAUX}.

In order to compactly represent a set of problems using different semantics for the same set of
formulae, multiple logic specifications can be put in a {\em problem generator} file, with 
multiple corresponding {\tt Status} values in the problem header.
These will be distributed in the {\tt Generators} directory of the TPTP problem library.
The TPTP4X utility will expand such files to multiple individual files with a single logic
specification and corresponding {\tt Status} value.
Selected individual files will be in the {\tt Problems} directory of the TPTP.

At the time of writing there is already one tool chain that can read, manipulate, and reason
over problems written in the TXN and THN languages, provided in the Leo-III framework \cite{SB21}.
Leo-III's parser \cite{Ste21}, also available as a stand-alone parsing library in Scala, can 
read both languages.
Problems in non-classical logic (including modal logic) are translated to THF using
a shallow embedding \cite{BP13,GSB17,GS18}, and reasoning proceeds using Leo-III's THF 
capabilities.
A generalization of the modal logic embedding procedure is available as and extensible library and 
executable \cite{Ste22-LE}, allowing any TPTP-compliant higher-order ATP system to be used
as the backend in this tool chain.
A tool to check the sanity of logic specifications for modal logics is available~\cite{Ste22-TU}.

%--------------------------------------------------------------------------------------------------
\section{Conclusion}
\label{Conclusion}

This paper has described the new TPTP languages, TXN and THN, for writing problems and solutions 
in non-classical logics.
TXN and THN employ new syntactic constructs to express non-classical logic connectives. 
The paper has additionally described a new type of annotated formula that is used to specify the 
logic to be used when reasoning. 
The use and flexibility of the proposed languages have been exemplarily demonstrated with modal 
logic problems.
The proposed syntax is quite general and unrestrictive, and makes no a priori statement about
semantics. 
Rather, the syntax provides a template that can be used with a logic defined in the TPTP, and
also in an ad-hoc way by users who would like to have a TPTP-oriented input syntax for their 
specialized context. 
While doing so, users benefit from the available TPTP infrastructure, parsers, translation tools 
and reporting standards.

Alongside the new syntax, the TPTP infrastructure is being extended to define and support various 
non-classical logics.
The TPTP technical manual 
is being updated to document the defined symbols used in these logics -- the connectives and 
their parameters, and the various components of their logic specification.
In conjunction with the technical manual, a suite of {\em logic files} that provide semi-formal 
machine-readable information about the non-classical logics defined in the TPTP is being 
developed.
The logic files will contain information such as logics' syntax, semantic and proof-theoretic 
parameterization, etc. 
This is ``work in progress'', which can be seen in the {\tt Logics} directory of the project
repository 
\href{https://github.com/TPTPWorld/NonClassicalLogic}{\texttt{https://github.com/TPTPWorld/NonClassicalLogic}}.

\paragraph{Further work.}
In the immediate future more non-classical logics will be standardised in the TPTP, and problems
in all the defined logics will be added to the TPTP problem library. 
As soon as an adequate number of problems and TPTP-compatible ATP systems are available for a
specific non-classical logic, a division for that logic will be added to CASC \cite{Sut16}. 
This will foster robust ATP system development for non-classical logic.
    
The TPTP syntax provides a very general framework for automated reasoning in expressive 
formalisms not yet addressed by this work, e.g., in knowledge representation it is often necessary 
to flexibly combine multiple logics to capture the different information dimensions 
\cite{CC+08-ALS,CC20}.
Typical examples include, e.g., combinations of temporal logic with (multi-agent) epistemic 
logics, and deontic logic with action languages.
Standard notions for systematically deriving combined logics from constituent logics in the 
context of normal modal logics are, among others, fusions \cite{Tho84} and fibrings \cite{Gab98}.
In the context of the TPTP syntax, it is intriguing to consider supporting fusions or fibrings 
by simply providing multiple logic specifications, yielding a very expressive and flexible 
representation for domain-specific logics.

%--------------------------------------------------------------------------------------------------
\bibliographystyle{splncs04}
\bibliography{Bibliography}
%--------------------------------------------------------------------------------------------------
\newpage
\appendix
\section{The TXN and THN Grammar}
\label{Grammar} 

\vspace*{1em}
{
\scriptsize
\begin{verbatim}
<thf_unitary_formula>  ::= <thf_quantified_formula> | <thf_atomic_formula> |
                           <variable> | (<thf_logic_formula>)
<thf_atomic_formula>   ::= <thf_plain_atomic> | <thf_defined_atomic> |
                           <thf_system_atomic> | <thf_fof_function>
<thf_defined_atomic>   ::= <defined_constant> | <thf_conditional> | <thf_let> |
                           <thf_defined_term> | (<thf_conn_term>) |
                           <tnc_connective>

<tff_unitary_formula>  ::= <tff_quantified_formula> | <tff_atomic_formula> |
                           <tfx_unitary_formula> | (<tff_logic_formula>)
<tff_atomic_formula>   ::= <tff_plain_atomic> | <tff_defined_atomic> |
                           <tff_system_atomic>
<tff_defined_atomic>   ::= <tff_defined_plain>
<tff_defined_plain>    ::= <defined_constant> |
                           <defined_functor>(<tff_arguments>) |
                           <tfx_conditional> | <tfx_let> | <tfx_tnc_atom>
<tfx_tnc_atom>         ::= <tnc_connective>(<tff_arguments>)
<tff_arguments>        ::= <tff_term> | <tff_term>,<tff_arguments>
<tff_term>             ::= <tff_logic_formula> | <defined_term> | <tfx_tuple> |
                           <tnc_key_pair>


<tnc_connective>       ::= <tnc_short_connective> | <tnc_long_connective>
<tnc_short_connective> ::= [.] | <less_sign>.<arrow> | <slash>.<backslash> | 
                           [<tnc_index>] | <less_sign><tnc_index><arrow> | 
                           <slash><tnc_index><backslash>
<tnc_long_connective>  ::= {<tnc_connective_name>} |
                           {<tnc_connective_name>(<tnc_parameter_list>)}
<tnc_connective_name>  ::= <def_or_sys_constant>
<tnc_parameter_list>   ::= <tnc_parameter> |
                           <tnc_parameter>,<tnc_parameter_list>
<tnc_parameter>        ::= <tnc_index> | <tnc_key_pair>
<tnc_index>            ::= <hash><tff_unitary_term>
<tnc_key_pair>         ::= <def_or_sys_constant> <assignment>
                           <tff_unitary_term>

%----Non-classical logic semantic specifications
<logic_defn_rule>      ::= <logic_defn_LHS> <identical> <logic_defn_RHS>
<logic_defn_LHS>       ::= <defined_constant>
<logic_defn_LHS>       :== $constants | $consequence | $modalities
<logic_defn_RHS>       ::= <defined_constant> | <tfx_tuple>
<logic_defn_RHS>       :== $rigid | $flexible |
                           $constant | $varying | $cumulative | $decreasing |
                           $modal_system_* | $modal_axiom_*
\end{verbatim}
}
%--------------------------------------------------------------------------------------------------
\newpage
\section{Detailed Example Problem}
\label{DetailedProblem}

The problem discussed in Section~\ref{ApplicationExample} is formalized in TXN as follows:
{
\scriptsize
\begin{verbatim}
    tff(tim,logic, 
        $modal == [
          $constants == $rigid,
          $quantification == $constant,
          $modalities == $modal_system_S5 ] ).
         
    % -- Type declarations
    tff(odd_decl, type, odd: $int > $o).
    tff(scMemberCount_decl, type, scMemberCount: $int).

    % -- Axiomatization of equality
    tff(eq_decl,  type,  eq: ($int*$int)>$o).
    tff(eq_refl,  axiom, ! [X:$int]: eq(X,X) ).
    tff(eq_sym,   axiom, ! [X:$int,Y:$int]: (eq(X,Y) => eq(Y,X))).
    tff(eq_trans, axiom, ! [X:$int,Y:$int,Z:$int]: ((eq(X,Y) & eq(Y,Z)) => eq(X,Z))).
    tff(eq_sub_1, axiom, ! [X:$int,Y:$int]: ((eq(X,Y) & odd(X)) => odd(Y))).
    
    % -- Problem axioms
    tff(four_members,hypothesis, eq(scMemberCount,4) ).
    tff(four_not_odd,hypothesis, ~ odd(4) ).
    tff(agreed_rule, hypothesis, {$necessary}(odd(scMemberCount)) ).
\end{verbatim}
}
Leo-III 1.6.7, internally reducing this problem to a classical HOL problem using the logic 
embedding tool \cite{Ste22-LE}, gives the following output:
{
\scriptsize
\begin{verbatim}
    > leo3 puzzle.tim.p
    % Axioms used in derivation (5): eq_sub_1, mrel_reflexive, agreed_rule, four_members, 
      four_not_odd
    % No. of inferences in proof: 22
    % SZS status Unsatisfiable for puzzle.tim.p : 1128 ms resp. 629 ms w/o parsing
\end{verbatim}
}
The axiom \texttt{mrel\_reflexive} does not exist in the original problem file but is introduced 
via the translation of \verb|$modal| to HOL. 
As can be seen, the axioms are already inconsistent in any modal logic system that includes axiom 
scheme $T$ (reflexivity).
%--------------------------------------------------------------------------------------------------

\end{document}